\documentclass{article}

\usepackage{PRIMEarxiv}

\usepackage[utf8]{inputenc} 
\usepackage[T1]{fontenc}    
\usepackage[colorlinks=true, linkcolor=black, citecolor=black, urlcolor=black]{hyperref}
\usepackage{url}            
\usepackage{booktabs}       
\usepackage{amsfonts}       
\usepackage{nicefrac}       
\usepackage{microtype}      
\usepackage{lipsum}
\usepackage{fancyhdr}       
\usepackage{graphicx}       
\graphicspath{{media/}}     

\usepackage{array}
\usepackage{geometry}
\usepackage{booktabs}
\usepackage{graphicx}
\usepackage{xcolor}
\usepackage{multirow}
\usepackage{rotating} 
\usepackage{pdfpages}
\usepackage{pdflscape}

\title{

Evaluation of Real-Time Preprocessing Methods in AI-Based ECG Signal Analysis

\thanks{\textit{2025 IEEE World AI IoT Congress (AIIoT) | 979-8-3315-2508-8/25/\$31.00 ©2025 IEEE | DOI: 10.1109/AIIoT65859.2025.11105222}
}}

\author{
  Jasmin Freudenberg, Kai Hahn, Christian Weber \\
  Medical Informatics and Graph-based Systems\\
  University of Siegen\\
  Am Eichenhang 50, 57068 Siegen, Germany\\
  \texttt{\{Jasmin Freudenberg\}jasmin.freudenberg@uni-siegen.de} \\
   \And
  Madjid Fathi \\
  Institute for Knowledge-Based Systems and Knowledge Management \\
  University of Siegen\\
  Hoelderlinstrasse 3, 57068 Siegen, Germany\\
  57068 Siegen, Germany
}

\begin{document}
\maketitle

\begin{abstract}
The increasing popularity of portable ECG systems and the growing demand for privacy-compliant, energy-efficient real-time analysis require new approaches to signal processing at the point of data acquisition. In this context, the edge domain is acquiring increasing importance, as it not only reduces latency times, but also enables an increased level of data security. The FACE project aims to develop an innovative machine learning solution for analysing long-term electrocardiograms that synergistically combines the strengths of edge and cloud computing. In this thesis, various pre-processing steps of ECG signals are analysed with regard to their applicability in the project. The selection of suitable methods in the edge area is based in particular on criteria such as energy efficiency, processing capability and real-time capability. 
\end{abstract}


\section{Introduction}
Cardiovascular diseases are currently recognised as the leading cause of mortality on a global scale. Therefore, the implementation of continuous cardiological care is a vital approach to promote a healthy lifestyle. One proposed method of achieving this objective is to monitor electrocardiogram signals in daily life. The electrocardiogram (ECG) is an effective tool for assessing cardiac health outside the hospital setting and facilitates long-term monitoring, a crucial aspect in the diagnosis of cardiovascular diseases. 
However, a manual analysis of such a large volume of biomedical signals by physicians is extremely labour-intensive and time-consuming \cite{Chen2024}.
Moreover, a 2020 review revealed that medical staff in cardiology, at all training levels, attained a median score of 54\% accuracy in the interpretation of test ECGs. Notably, even cardiologists with extensive experience attained a maximum score of 75\% accuracy \cite{Cook2020}.
As a result, data-driven methodologies for ECG monitoring have been developed to such a degree that they are now capable of attaining a classification performance that is comparable to that of a cardiologist \cite{Chen2024}.\\
Conventionally, the training and testing of such AI models has been conducted on cloud servers that are equipped with sufficient computational resources, thereby providing a viable solution for ECG classification tasks. Nevertheless, challenges persist in relation to privacy risks, response time and processing costs.
Edge computing offers a promising solution to the aforementioned challenges, functioning as a complement to traditional cloud computing. It exhibits lower power consumption and latency, in addition to the capacity to allocate resources and outsource tasks \cite{Chen2024}.\\
The objective of the FACE project is therefore to design suitable software and hardware environments for the development and optimisation of edge and cloud-based AI models for ECG analysis. The aim of this paper is to provide a comparison of appropriate pre-processing methodologies in the context of edge computing. 
\newpage
\includepdf[pages=-, fitpaper=true, pagecommand={\label{tab:techniques}}]{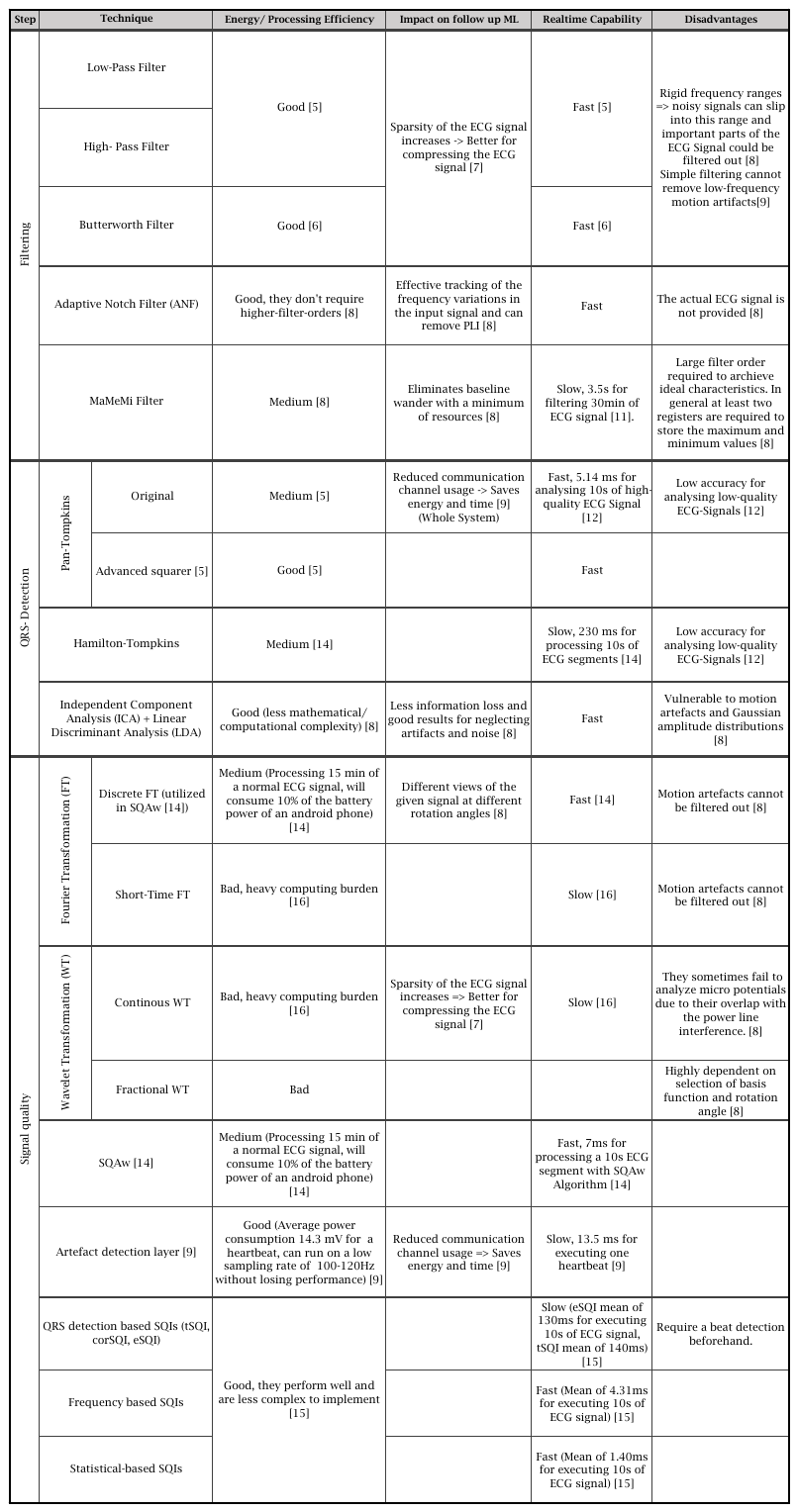} 

\section{Pre-processing steps on the Edge Model}
A comprehensive literature review was conducted to systematically compare the power consumption and processing efficiency, the impact on follow-up methods, the real-time capability, and the disadvantages of state-of-the-art preprocessing techniques. The outcomes of this review are outlined in table \ref{tab:techniques} below.
\subsection{Filtering}
The ECG signal is typically subject to bandpass filtration to eliminate noise artefacts. The most common frequency range used for bandpass filtering is 0.1–100 Hz \cite{KaplanBerkaya2018}.
A multitude of bandpass filters exist, each exhibiting a unique operational characteristics. An Butterworth filter is an easy to apply version of it \cite{CarrickUtsi2017}. 
In the course of the study undertaken by Nagatomo et al. in 2024, low- and high pass filters were analysed as subitems of the Pan-Tompkins algorithm. 
The study concluded that high-pass and low-pass filtering can be simplified to the point where only adders are required. Furthermore, it was determined that the power consumption of the adders can be neglected in comparison to that of the multipliers \cite{Nagatomo2024}. Consequently, the filters were found to be energy-efficient and highly processing-efficient.\\
The study by Septian et al. (2024) analysed Electromyography noise suppression with a low complexity algorithm, for which the Butterworth Band Pass Filter algorithm was optimised into Butterworth BPF D-II. The processing time and memory usage requirements of Butterworth BPF D-I and the proposed method were then compared. The outcomes of this study demonstrated a reduction in processing time and memory usage by 0.03s and 0.024KB, respectively \cite{Septian2024}.
These findings also emphasise the efficient processing capabilities.\\
In the study conducted by Khorasani (2019), an analysis was undertaken of the structure and features of ECG signals subsequent to the implementation of pre-processing steps. The investigation revealed that signal sparsity is increased by pre-processing, thus facilitating enhanced compression. Furthermore, it was determined that compressing the wavelet-transformed or filtered signal with greater sparsity content results in reduced compression error and superior reconstruction \cite{Khorasani2019}.
It was also established that low- and high-pass filters are suitable for real-time applications. This is due to the fact that they are both rapid and exhibit minimal power consumption and memory usage \cite{Nagatomo2024,Septian2024}.\\
A potential disadvantage of digital bandpass filtering is that its performance is affected by the time-varying nature of the ECG signal. This is due to the fact that the amplitude, shape and frequency of an ECG signal change over time due to movement, breathing or cardiac activity. Such variations can occur during physical activities or abruptly, as in cases of cardiac arrhythmias. A significant challenge arises when critical components within the ECG signal are inadvertently filtered out, a consequence of fluctuations in signal frequency over extended periods. Moreover, the capacity of these filters to effectively suppress interference signals is significantly diminished when such signals enter the frequency range of the filter. Additionally, in certain instances, micro-potentials may not be analysed if they coincide with power line interference (PLI) \cite{Gupta2022}. Moreover simple filtering cannot remove low-frequency motion artifacts \cite{Demirel2022}.\\
Kaplan et al. (2018) compared the accuracy of band-pass, low-pass and high-pass filters by means of a tabular representation. The accuracy of band-pass filters was found to range from 94.8\% to 100\%, while that of low-pass filters was found to range from 80\% to 99.6\%. The accuracy of high-pass filters was found to range from 88.84\% to 100\% \cite{KaplanBerkaya2018}.\\
Gupta et al. conducted 2022 a comprehensive review of the extant literature on pre-processing ECG signals, with the aim of composing a novel technique that would effectively remove interference and noise, while maintaining an improvement in the classification of ECG signals. In this review, they evaluated a range of methodologies, including an adaptive notch filter (ANF) or a maximum-mean-minimum (MaMeMi) filter.\\
The ANF is widely regarded as a benchmark technique for processing and analysing ECG signals. A distinguishing feature of the ANF is its ability to function without a higher filter order. In fact the structure is composed of a cascade of second-order sections. This is in contrast to ordinary analog and digital filters, which require higher filter orders. Moreover, ANF demonstrates an inherent adaptability and the capacity to eliminate PLI, thereby facilitating effective tracking of frequency variations in the input signal \cite{Gupta2022}.\\ 
Adaptability in this context signifies the ability of the filter parameters to be adjusted in accordance with the characteristics of the input signal. It continuously updates its coefficients based on the input signal, allowing it to adapt to changing conditions and effectively remove noise while preserving the desired signal components \cite{10827499}. 
In order to achieve this objective, a combination of finite impulse response (FIR) and infinite impulse response (IIR) is utilised to facilitate the real-time update of the latest coefficients \cite{Gupta2022}. 
As both low- and highpass filtering are combined in notch filters \cite{KaplanBerkaya2018}, it may be concluded that ANFs are also capable of real-time operation.\\
However, it should be noted that one disadvantage of the ANF is that it will not provide the actual ECG signal. Additionally the Independent Component Analysis (ICA) Method proposed by \cite{Gupta2022} outperformes the ANF at the preprocessing stage.\\
The MaMeMi filter is a combination of two distinct mathematical processes: a butterworth high-pass filter and an adaptive empirical mode decomposition method. The former serves to filter out high and low frequency noise components from an input ECG signal, while the latter decomposes the signal into a series of intrinsic mode functions. The collaborative effect of these processes is to facilitate the effective management of unpredictable changes in ECG waveforms \cite{Sheetal2018}.\\
The performance of the filter is contingent upon the sampling frequency and the calibration of the filter coefficients. In addition, it is necessary to have at least two registers in order to accommodate the maximum and minimum values. One advantage of this method is that it can eliminate baseline wander with a minimum of resources \cite{Gupta2022}.
Due to the two memory registers required, the processing efficiency is rated as average.\\
The real-time capability of MaMeMi is limited, since the comparison of elapsed time for filtering 30 minutes of ECG signal is 3.5 seconds \cite{Sheetal2018}.
\subsection{QRS Detection}
\label{subsubsec: QRS}
The Pan-Tompkins algorithm is the most common QRS detection algorithm \cite{KaplanBerkaya2018}. It is comprised of six distinct steps for the detection of the QRS complex:
\begin{enumerate}
    \item Low-pass and high-pass filtering
    \item Differentiation
    \item Squaring
    \item Moving window averaging
    \item Adaptive threshold detection \cite{Nagatomo2024}
\end{enumerate}
Nagatomo et al. (2024) determined that the squaring process is the sole step that necessitates the utilisation of a multiplier. The remaining steps in the algorithm can be executed through the use of adders alone. As previously referenced, the power consumption of the adders can be disregarded when contrasted with that of the multipliers. As a result the multiplier was substituted with a power-efficient squarer, thereby achieving a substantial reduction in the total power requirement for the Pan-Tompkins algorithm \cite{Nagatomo2024}.\\ Consequently, the power consumption and processing efficiency of the "normal" Pan-Tompkins is medium and that of the optimised one is good.\\
In the study conducted by Demirel et al. (2022), the Pan-Tompkins approach was utilised to identify R-peaks. The detected R-peaks were then employed to segment the heartbeat and subsequently calculate the heartbeat rate variability (HRV) between three consecutive beats. Furthermore, 20 regular beats were averaged and aligned according to their R-peaks, after which the correlation was calculated. These indicators were used to classify whether a heartbeat is abnormal or normal. Then the energy consumption of transmitting the ECG signal to the fog/cloud node of the proposed model was examined. It was determined that the elimination of normal and noisy beats prior to transmission can reduce the load on the fog/cloud server and minimise communication channel usage. This approach has the potential to conserve energy and reduce time expenditure \cite{Demirel2022}.\\
Liu et al. (2017) investigated the performance as well as the noise responses of three widely used Pan-Tompkins-based QRS complex detection algorithms. The study revealed that the original version of the Pan-Tompkins method was the fastest, with an average time cost of 5.14 milliseconds for analysing 10 seconds of high-quality ECG signal and an average time cost of 5.67 milliseconds for 10 seconds of low-quality ECG signal. Moreover all three Pan-Tompkins algorithms demonstrated high detection accuracy's on the high-quality ECG database, with percentages exceeding 99\%. However, it should be noted that the detection accuracy's reported for the low-quality ECG database were comparatively low, with the original Pan-Tompkins method achieving the lowest accuracy of 74.49\% \cite{Liu2017}.\\
The Hamilton-Tompkins algorithm represents an enhancement to the original version proposed by Pan and Tompkins in 1985, which involves the utilisation of a patient-specific threshold for QRS complex detection \cite{Rahman2012AssessmentOR}.
Satija's (2017) study proposed a lightweight, real-time, signal quality-aware (SQA) technique for classifying electrocardiogram (ECG) signals into acceptable or unacceptable noise classes. To this end, the performance respectively the computational load of four different SQA methods was analysed. The Hamilton-Tompkins-based QRS detector and template matching method exhibited an average computational time of 230 milliseconds for the processing of a 10-second ECG segment. The sensitivity for detecting clean signal was  78.94\% and for detecting noisy signal 69.36\% \cite{Satija2017}.\\
As this is a long time compared to the proposed method in the paper, the processing efficiency is rated as medium and the realtime capability is rated as low.\\
In the study by Gupta et al. (2022), a combined ICA and linear discriminant analysis (LDA) method was selected to detect the R-peak in ECG signals. The rationale behind this choice was that the ICA has been demonstrated to exhibit superior performance in comparison to seven higher-order filters (both digital and analogue) with regard to information loss and mathematical and computational complexity \cite{Gupta2022}.
ICA can be defined as a non-linear method for reducing the dimensions of overlapping signals. In this method, the signal is decomposed into a series of distinct, independent additive subcomponents. While the method has been shown to be effective in the elimination of artefacts and noises, it is susceptible to the presence of motion artefacts and is ineffective when it comes to Gaussian amplitude distributions \cite{Gupta2022}.\\
The LDA was utilised due to its capacity to minimise false detections by maximising the distance between data classes and minimising the variance between them. Furthermore, the dimensional costs are reduced by utilising the LDA. 
A further comparison was made between the performance of a wavelet and an SVM classifier, as well as that of ICA and LDA techniques. The findings revealed that the wavelet and SVM classifier exhibited a sensitivity of 82.75\%, while the ICA and LDA technique demonstrated 99.92\% sensitivity \cite{Gupta2022}.\\ 
The real-time capability is estimated as fast, since the mathematical and computational complexity is described as low.
\subsection{Signal Quality}
\label{subsubsec:SG}
ECG signals are susceptible to interference from various sources, including poor electrode quality, respiration and current flow in the system cables. The interference's can be classified as baseline wander (BLW), motion artefacts (MA), a additive white Gaussian noise (AGN), abrupt changes (AC) or power line interference (PLI). Consequently, these disturbances have the potential to obscure diagnostic information crucial for the accurate assessment of cardiac disease during signal acquisition \cite{Gupta2022, Kuetche2023, Satija2017}.
There are many filter strategies proposed in the literature. This study focuses on fourier- and wavelet tranform, a signal quality-aware technique (SQAw) proposed by \cite{Satija2017}, an artifact detection layer according to \cite{Demirel2022} and QRS detector based signal quality indices (SQIs) presented in \cite{Kuetche2023} as well as Support Vector Machines (SVM), Random Forest (RF).\\
In 2023, Han et al. presented an edge-based model for real-time ECG classification and arrhythmia detection, it uses continuous wavelet transform (CWT) and short-time fourier transform (STFT) to convert 1D ECG signals into a 2D heart map, which is then processed by a 2D convolutional neural network (CNN)- long short-term memory (LSTM) network. The evaluation process revealed that utilising a combination of STFT and CWT results in a heavy computing burden.
The CWT is used to analyse signals in the time domain. It visualises the frequency components and preserves the time information at the same time \cite{Han2023}.\\
It has been demonstrated that both bandpass filtering and wavelet transformation encounter difficulties when attempting to analyse ECG signals, due to the time-varying characteristics of it. In certain instances, these methodologies have been observed to encounter deficiencies in the analysis of micro potentials, a phenomenon that can be attributed to their susceptibility to interference from PLI. It is therefore imperative to select a proper basis function with scale \cite{Gupta2022}.\\
As mentioned above Satija et al. (2017) proposed a signal quality-aware technique (SQAw) for classifying ECG-signals into acceptable or unacceptable noise classes. The ECG signal quality assessment (SQA) is conducted through the implementation of a discrete Fourier transform (DFT)-based filtering process, the identification of turning points (TPs), and the establishment of decision rules. The detection of BLW and AC was facilitated by the DFT.
The proposed method was implemented on an Android phone to analyse power and time consumption of the whole system. The investigation revealed that the analysis of 15 minutes of a standard ECG signal results in the consumption of approximately 10\% of the Android battery's capacity.  Furthermore, it was determined that the analysis of 15 minutes of a standard ECG signal required a processing duration of 7 milliseconds. The whole technique had an sensitivity value of 95.56\% for clean signal and 97.85\% for noisy signals \cite{Satija2017}.\\
In the noise detection layer of the proposed model by \cite{Demirel2022}, a rule-based algorithm is utilized to classify signals into acceptable and unacceptable groups. This is achieved by a median check, calculating the moving standard deviation and comparing the mean value of the waveform with a predefined threshold. The noise detection process demonstrated satisfactory energy efficiency, as it was capable of operating at a low sampling rate of 100–120 Hz without any discernible decline in performance. The energy requirements are 193.05$\mu$ J, the execution time is 13.5 ms, and the average power consumption is 14.3 mV for a heartbeat. Furthermore, the model is compatible with devices that have a minimum random-access memory (RAM) of 32 kilobytes \cite{Demirel2022}.\\
It was moreover discovered that the rule-based decision method previously outlined is more effective than classical machine learning algorithms, such as Random Forest (RF) or Support Vector Machine (SVM), in terms of power efficiency. Furthermore, the rule-based decision method exhibited equivalent classification performance in detecting noise, with an sensitivity of 99.2\% for noisy signals and 99.4\% for clean signals \cite{Demirel2022}.\\
In 2023, Kuetsche et al. conducted a comprehensive analysis of the most commonly used signal quality indices (SQIs) in signal quality assessment (SQA) systems. The study encompassed a systematic comparison of the robustness and complexity of these indices in terms of computational speed. This analysis was conducted by extracting the time required for analysing 10-second ECG signals of a total of 39 SQIs. The SQIs included in the study varied in their underlying mechanism, with 7 based on QRS detectors, 9 frequency-based, 7 non-linear, and 16 statistical \cite{Kuetche2023}.\\ 
The study revealed that frequency-based SQIs and statistical-based SQIs are the fastest, with a mean of 4.31 ms and 1.40 ms, respectively. Non-linear SQIs were the most complex, with an execution time of 126 milliseconds. Conversely, beat detector-based SQIs (corSQI, tSQI, and eSQI) were the most efficient, due to their ease of implementation and optimal performance. The lowest execution times among the QRS detection-based SQIs were observed in eSQI, with an mean time of 130ms and corSQI, with a mean time of 140ms. The corSQI and tSQI demonstrated the highest performance for all noise, with an Area Under the Receiver Operating Characteristic curve (AUC) of 0.96. All in all a combination of several SQIs was recommended \cite{Kuetche2023}.
\section{Results}
This paper provides a synopsis of the inaugural scientific outcomes derived from the FACE project, with a particular focus on the pre-processing of ECG signals. A review of different state-of-the-art pre-processing methods was conducted, which revealed that standard bandpass filters are well designed for real-time tasks; however, they have a common disadvantage in that they have fixed frequency limits. This is problematic when dealing with naturally time-varying ECG signals. Adaptive filters are a good solution for that, since frequency variations are effectively tracked \cite{Gupta2022}. Nevertheless, it is essential to determine the time requirements of the ANF filtration processes.
The MaMeMi filter can eliminate baseline wandering with minimal resources, but takes longer to perform compared to other preprocessing techniques.\\
The original Pan-Tompkins algorithm exhibits suboptimal energy efficiency in comparison to the approach proposed by \cite{Nagatomo2024}. However, both algorithms are characterised by a high processing speed, even if the exact time required was not specified in the evaluation of the advanced Pan-Tompkins algorithm. In addition, its performance should be tested on poor quality ECG signals. In comparison with the original, the Hamilton-Tompkins algorithm is slow, and both algorithms demonstrate suboptimal performance on low-quality ECG signals.\\
The ICA has been demonstrated to exhibit low computational and mathematical complexity, in addition to reduced information loss when compared with alternative preprocessing steps. LDA on the other hand cuts down the dimensional costs. Nevertheless, it is susceptible to motion and Gaussian artefacts. Furthermore, it should be noted that a concrete inference time test was not performed on the combined ICA + LDA method.\\
The different Fourier- and Wavelet Transformation types are mostly computational intensive and therefore not that realtime capable, since they consume much battery power and computing power. The benefit of utilizing FT is that the signal is shown in different rotation angles, however a disadvantage is that it isn't capable of removing motion artefacts. The utilization of a WT results in a easier compression of the signal afterwards, but sometimes micro potentials cannot be analysed due to overlaps with PLIs.
SQAw is a rapid solution for signal quality analysis; however, further research is required to address the high battery consumption.\\
The artefact detection layer from \cite{Demirel2022} is power efficient; moreover, its utilisation reduces the communication channel usage and therefore saves time and power by filtering out normal and noisy heartbeats before transmission. However, its execution time is comparatively longer than that of SQAw and various SQIs. Nevertheless, the proposed model demonstrates the greatest sensitivity, specificity, PVV and accuracy of all the models under consideration.\\
The most efficient SQIs identified by \cite{Kuetche2023} were tSQI, corSQI and eSQI, as well as frequency and statistical-based SQIs. These SQIs were chosen because they perform well and are less complex to implement. Nevertheless, the SQAw Method of \cite{Satija2017} was found to be significantly faster than the QRS detection-based SQIs, yet slower than the frequency and statistical-based SQIs.\\
A significant limitation of this study is the absence of quantitative metrics for a few algorithms, as not all original publications report them. This limitation restricts the capacity for objective comparison and evaluation of the presented approaches.

\section{Conclusion}
All in all adaptive filters, such as the ANF, have proven to be an effective solution for problems with fixed frequency ranges in bandpass filtering, while also being well suited for real-time tasks. In addition to ICA+LDA, the extended Pan-Tompkins method proposed by \cite{Nagatomo2024} is a promising, real-time capable solution for extracting the QRS complex with low power consumption. The ICA approach is characterised by the fact that it is not computationally intensive and mathematically complex. Another advantage is that it is only vulnerable for low-quality signals caused by movement or Gaussian artefacts. In terms of signal quality assessment, SQAw and the artefact detection layer of \cite{Demirel2022} are considered the most promising solutions. Since SQAw is demonstrably faster than SQIs based on QRS detection, but has a similar high sensitivity. Nevertheless, the power consumption of the artefact detection layer is better than that of SQAw, as it can operate at a low sampling rate. Although the artefact detection layer processes ECG signals more slowly than the SQIs and SQAw, its performance is superior.\\
Future work within the FACE project will focus on evaluating the most promising techniques using a common reference dataset and standardised hardware such as the NVIDIA Jetson AGX Orin. This approach will enable a comprehensive comparison. 

\section{Funding}
This ongoing research is funded by the German Federal Ministry of Economic Affairs and Climate Action (BMWK) and European Union "NextGenerationEU" (13IPC031F). The project is being realised in collaboration with the following partners: GETEMED Medizin- und Informationstechnik AG, Charité (University Medicine Berlin), BIOTRONIK Vertriebs GmbH \& Co. KG, SEMDATEX GmbH, Evangelisches Diakonissenhaus Berlin Teltow
Lehnin Stiftung bürgerlichen Rechts and the Professorship for Medical Informatic and
Microsystems Design at the University of Siegen.

\bibliographystyle{unsrt}  
\bibliography{templateArXiv}

\begin{thebibliography}{10}

\bibitem{Chen2024}
Jiarong Chen, Xianbin Zhang, Lin Xu, Victor Hugo~C. de~Albuquerque, and Wanqing
  Wu.
\newblock Implementing the confidence constraint cloud-edge collaborative
  computing strategy for ultra-efficient arrhythmia monitoring.
\newblock {\em Applied Soft Computing}, 154:111402, March 2024.
\newblock DOI: 10.1016/j.asoc.2024.111402.

\bibitem{Cook2020}
David~A. Cook, So-Young Oh, and Martin~V. Pusic.
\newblock Accuracy of physicians’ electrocardiogram interpretations: A
  systematic review and meta-analysis.
\newblock {\em JAMA Internal Medicine}, 180(11):1461, November 2020.
\newblock DOI: 10.1001/jamainternmed.2020.3989.

\bibitem{KaplanBerkaya2018}
Selcan Kaplan~Berkaya, Alper~Kursat Uysal, Efnan Sora~Gunal, Semih Ergin,
  Serkan Gunal, and M.~Bilginer Gulmezoglu.
\newblock A survey on ecg analysis.
\newblock {\em Biomedical Signal Processing and Control}, 43:216--235, May
  2018.
\newblock DOI: 10.1016/j.bspc.2018.03.003.

\bibitem{CarrickUtsi2017}
Erica~Carrick Utsi.
\newblock {\em Ground penetrating radar: theory and practice}.
\newblock Butterworth-Heinemann, 2017.

\bibitem{Nagatomo2024}
Taiki Nagatomo and Toshinori Sato.
\newblock Cost-effective alternatives for squarer in pan-tompkins algorithm.
\newblock In {\em Proceedings of the 2024 6th International Electronics
  Communication Conference}, IECC 2024, pages 31--36. ACM, July 2024.
\newblock DOI: 10.1145/3686625.3686631.

\bibitem{Septian2024}
Belen Septian.
\newblock Noise suppression of ecg signal using optimized digital butterworth
  bandpass filter.
\newblock {\em The Indonesian Journal of Computer Science}, 13(4), August 2024.
\newblock DOI: 10.33022/ijcs.v13i4.4312.

\bibitem{Khorasani2019}
S.~Monem Khorasani, G.A. Hodtani, and M.~Molavi Kakhki.
\newblock Investigation and comparison of ecg signal sparsity and features
  variations due to pre-processing steps.
\newblock {\em Biomedical Signal Processing and Control}, 49:87--95, March
  2019.
\newblock DOI: 10.1016/j.bspc.2018.11.004.

\bibitem{Gupta2022}
Varun Gupta, Monika Mittal, and Vikas Mittal.
\newblock A simplistic and novel technique for ecg signal pre-processing.
\newblock {\em IETE Journal of Research}, 70(1):815--826, October 2022.
\newblock DOI: 10.1080/03772063.2022.2135622.

\bibitem{Demirel2022}
Berken~Utku Demirel, Islam~Abdelsalam Bayoumy, and Mohammad Abdullah~Al
  Faruque.
\newblock Energy-efficient real-time heart monitoring on edge–fog–cloud
  internet of medical things.
\newblock {\em IEEE Internet of Things Journal}, 9(14):12472--12481, July 2022.
\newblock DOI: 10.1109/jiot.2021.3138516.

\bibitem{10827499}
Yingwen Yao and Yanbing Jiang.
\newblock A physiological signal denoising method using adaptive kalman filter.
\newblock In {\em 2024 7th International Conference on Pattern Recognition and
  Artificial Intelligence (PRAI)}, pages 968--972, 2024.
\newblock DOI: 10.1109/PRAI62207.2024.10827499.

\bibitem{Sheetal2018}
Anu Sheetal, Harjit Singh, and Anureet Kaur.
\newblock Qrs detection of ecg signal using hybrid derivative and mamemi filter
  by effectively eliminating the baseline wander.
\newblock {\em Analog Integrated Circuits and Signal Processing}, 98(1):1--9,
  June 2018.
\newblock DOI: 10.1007/s10470-018-1249-7.

\bibitem{Liu2017}
Feifei Liu, Shoushui Wei, Yibin Li, Xinge Jiang, Zhimin Zhang, Ling Zhang, and
  Chengyu Liu.
\newblock The accuracy on the common pan-tompkins based qrs detection methods
  through low-quality electrocardiogram database.
\newblock {\em Journal of Medical Imaging and Health Informatics},
  7(5):1039--1043, September 2017.
\newblock DOI: 10.1166/jmihi.2017.2134.

\bibitem{Rahman2012AssessmentOR}
Saeka Rahman and Mohammad~A. Rahman.
\newblock Assessment of reliability of hamilton-tompkins algorithm to ecg
  parameter detection.
\newblock 2012.
\newblock Last seen 10.03.25.

\bibitem{Satija2017}
Udit Satija, Barathram Ramkumar, and M.~Sabarimalai~Manikandan.
\newblock Real-time signal quality-aware ecg telemetry system for iot-based
  health care monitoring.
\newblock {\em IEEE Internet of Things Journal}, 4(3):815--823, 2017.
\newblock DOI: 10.1109/JIOT.2017.2670022.

\bibitem{Kuetche2023}
Fotsing Kuetche, Noura Alexendre, Ntsama~Eloundou Pascal, Welba Colince, and
  Simo Thierry.
\newblock Signal quality indices evaluation for robust ecg signal quality
  assessment systems.
\newblock {\em Biomedical Physics \&amp; Engineering Express}, 9(5):055016,
  August 2023.
\newblock DOI: 10.1088/2057-1976/ace9e0.

\bibitem{Han2023}
Yibo Han, Pu~Han, Bo~Yuan, Zheng Zhang, Lu~Liu, and John Panneerselvam.
\newblock Novel transformation deep learning model for electrocardiogram
  classification and arrhythmia detection using edge computing.
\newblock {\em Journal of Grid Computing}, 22(1), December 2023.
\newblock DOI: 10.1007/s10723-023-09717-3.

\end{thebibliography}

\end{document}